\documentclass[a4paper]{interact}

\usepackage{epstopdf}
\usepackage{graphicx}
\usepackage{subcaption}
\usepackage{placeins}
\usepackage{wrapfig}
\usepackage{pdflscape}
\usepackage[normalem]{ulem}
\usepackage{soul}
\usepackage{xcolor}

\AtBeginDocument{%
  }

\usepackage[numbers,sort&compress]{natbib}
\bibpunct[, ]{[}{]}{,}{n}{,}{,}

\theoremstyle{plain}

\theoremstyle{definition}

\theoremstyle{remark}

\begin{document}


\title{Causal Discovery and Counterfactual Reasoning to Optimize Persuasive Dialogue Policies}


\author{
\name{Donghuo Zeng\textsuperscript{1*}\thanks{*Both authors contributed equally to this research. Corresponding author is D.Zeng, xdo-zen@kddi.com}, Roberto Legaspi\textsuperscript{1*}\footnotemark[1], Yuewen Sun\textsuperscript{2}, Xinshuai Dong\textsuperscript{2}, Kazushi Ikeda\textsuperscript{1}, Peter Spirtes\textsuperscript{2}, and Kun Zhang\textsuperscript{2}}
\affil{\textsuperscript{1}Human-Centered AI Labs, KDDI Research, Inc., Saitama, Japan. \\ \textsuperscript{2}Department of Philosophy, Carnegie Mellon University, Pittsburgh, PA USA}
}

\maketitle

\begin{abstract}
Tailoring persuasive conversations to users leads to more effective persuasion. However, existing dialogue systems often struggle to adapt to dynamically evolving user states. This paper presents a novel method that leverages causal discovery and counterfactual reasoning for optimizing system persuasion capability and outcomes. We employ the Greedy Relaxation of the Sparsest Permutation (GRaSP) algorithm to identify causal relationships between user and system utterance strategies, treating user strategies as states and system strategies as actions. GRaSP identifies user strategies as causal factors influencing system responses, which inform Bidirectional Conditional Generative Adversarial Networks (BiCoGAN) in generating counterfactual utterances for the system.
Subsequently, we use the Dueling Double Deep Q-Network (D3QN) model to utilize counterfactual data to determine the best policy for selecting system utterances. Our experiments with the PersuasionForGood dataset show measurable improvements in persuasion outcomes using our approach over baseline methods. The observed increase in cumulative rewards and Q-values highlights the effectiveness of causal discovery in enhancing counterfactual reasoning and optimizing reinforcement learning policies for online dialogue systems.

\end{abstract}


\section{Introduction} \label{intro}
Persuasive conversations~\cite{prakken2006formal,torning2009persuasive,weietal2016post,yoshino2018dialogue} play a pivotal role in shaping user opinions, attitudes, and behaviors, particularly within the realms of marketing~\cite{braca2023developing}, health communication~\cite{reynolds2019psychological, dillard2005nature}, and social interventions~\cite{kelders2012persuasive}. The effectiveness of these conversations hinges on adapting communication strategies to account for individual differences among users, such as in terms of their personality traits~\cite{papangelis2022understanding}, current situation~\cite{lee2006situation},
and prior interactions~\cite{ribeiro2015influence}. Conventional systems for persuasive dialogues typically depend on fixed interaction policies known beforehand or rigid models that overlook the intricate and evolving aspects of the interactions, resulting in less effective persuasions.

Current research on persuasive dialogue systems has mainly concentrated on how predefined persuasive strategies influence potential outcomes~\cite{shi2020effects,tran2022ask}, which often neglects the dynamic and personalized adaptation required to suit individual users. For example, the study in~\cite{hirsh2012personalized} contends that the order of persuasive strategies aimed at specific demographics may not be significant, advocating instead for a consistent sequence of persuasive appeals. While these systems have achieved some success in strategy learning, they lack the flexibility to adapt to users' potentially ever-changing states, which can lead to unsatisfactory or undesirable persuasion outcomes.

\begin{figure}
    \centering
    \includegraphics[scale=0.55]{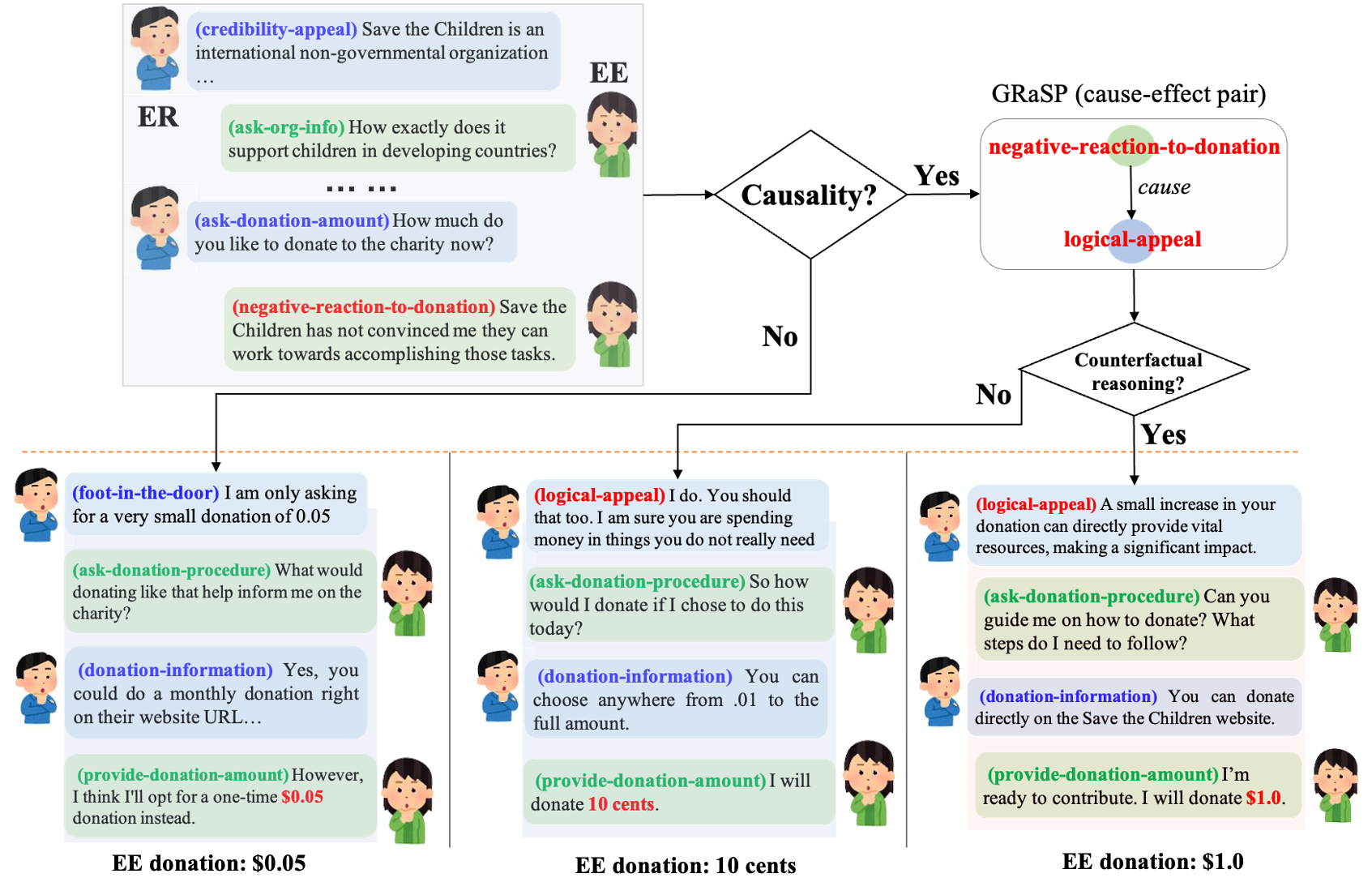}
    \caption{An example of a persuasive dialogue between the persuader (ER) and persuadee (EE), which was extracted from the PersuasionForGood dataset (ID:20180826-181951\_904\_live, donation: \$0.05).  
Each utterance of both ER and EE has been annotated in the dataset with a strategy. By identifying the causal relationship between EE's \textit{negative-reaction-to-donation} strategy and ER's \textit{logical-appeal} strategy, the persuader can replace the ground-truth \textit{foot-in-the-door} strategy in the PersuasionForGood dataset with the more effective \textit{logical-appeal} strategy. If counterfactual reasoning (CL) is applied, BiCoGAN generates the next action with \textit{logical-appeal} strategy; otherwise, it is selected from the ground truth with \textit{logical-appeal} strategy. The resulting donations are: ground truth (\$0.05), causality without CL (\$0.10), and causality with CL (\$1.00).}
    \label{fig:example_causal}
\end{figure}

To address the above challenges, we innovate from our work in~\cite{zeng2024counterfactual} to now focus on causal discovery guiding counterfactual reasoning. In contrast to existing methods that depends on already established strategies or to random selection of counterfactual actions, we use the  Greedy Relaxation of the Sparsest Permutation (GRaSP
)~\cite{lam2022greedy, zheng2024causal} algorithm to discover from the data the cause-effect relationships at the strategy level of the dialogue utterances. The discovered causal relationships is then leveraged by a Bidirectional Conditional Generative Adversarial Network (BiCoGAN)~\cite{jaiswal2019bidirectional,lu2020sample} to select intelligibly from a wide range of alternative scenarios when constructing the counterfactual dialogue dataset.

Specifically, our approach identifies the strategies dictating the persuadee's ongoing utterances and treats them as the causal factors. This enables us to leverage the strategies associated with these factors, i.e., the effect strategies, as counterfactual actions, which enable BiCoGAN to produce alternative utterances. 
Let us take for instance the dialogue sample in Fig.~\ref{fig:example_causal} between a persuader (ER) and persuadee (EE). In the dataset, the utterances of both ER and EE had been annotated with strategies. Our approach would identify which among the strategies of ER would result from the causal influence of EE's strategies. At the upper right side of Fig. 1~\ref{fig:example_causal}, we see that EE's \textit{negative-reaction-to-donation} strategy has been inferred to have caused ER to respond with an utterance that follows a \textit{logical-appeal} strategy. With this causal knowledge, the counterfactual reasoning part at the bottom right will take this information to provide an alternative answer. Had there been no causal discovery involved, as shown at the left of Fig. 1~\ref{fig:example_causal}, the ER strategy would take on the ground truth strategy in the data, while disregarding the causal structure that plausibly underlies the interactions of ER and EE. The ER adapting its response from a supposed \textit{foot-in-the-door} strategy to \textit{logical-appeal}. If we apply counterfactual reasoning using BiCoGAN to generate the next action with \textit{logical-appeal} strategy, it would increase the donation of \$0.05 to \$1.0; otherwise the next state with \textit{logical-appeal} strategy will be selected from the ground truth database, it will increase the donation of \$0.05 to \$0.1.


At the tail end of our framework, we optimize the selection of system utterances based on the counterfactual data using the Dueling Double Deep Q-Network (D3QN)~\cite{raghu2017deep}. By learning the best interaction protocols from the causality-navigated counterfactual data, the system is better able to dynamically adjust its responses. ultimately improving the quality and effectiveness of its persuasive policies. 

Our experimental results using the PersuasionForGood dataset~\cite{WangSKOYZY19} validate the ability of our proposed method to significantly improve the cumulative rewards and Q-values compared to our methods in~\cite{zeng2024counterfactual} as baseline. 
Our findings highlight how reinforcement learning policies for online persuasive system interactions can be improved by combining causal discovery and counterfactual reasoning.

To summarize, the contributions of this work are as follows:
\begin{itemize}
    \item We integrated causal discovery into the system response generation process using the GRaSP algorithm to identify causal relationships at the strategy level. 
    This allows us to generate counterfactual actions that are established on underlying causal relationships.
    \item The BiCoGAN model aided by the discovered strategy-level causal relationships generates effective counterfactual data. This generated data embodies the alternative and enhanced utterances that extend the original dialogues of the PersuasionForGood dataset.
    \item The D3QN model learns the optimal policy from the causal graph-based counterfactual data, which enables the system to dynamically adapt its responses conditioned by user states, which would increase the quality of persuasive conversations yielding desirable persuasion outcomes.  
\end{itemize}

\section{Related Works}\label{related}

\subsection{Persuasive Strategies}
Dialogue data from online discussions and social media that are annotated with persuasive strategies are often used to enhance argument-mining techniques~\cite{chakrabarty2019,wachsmuth2017computational,yang2019let,wei2016post} for building dialogue systems.
Although these studies have contributed valuable methodologies to realizing persuasion strategies, the exploration of efficient and automated methods for their effective application remains limited
In contrast to previous approaches,~\cite{WangSKOYZY19} developed the PersuasionForGood dataset, a crowdsourced collection specifically designed to model donation-focused persuasive scenarios in conversational formats. This dataset enables a structured analysis of persuasive strategies within realistic, dialogue-based contexts, enhancing research into effective persuasion techniques. The work in~\cite{shi2020effects} leveraged this dataset to develop a persuasive dialogue system applying an agenda-based method. However, this method relied on a predetermined sequence of persuasive strategies and did not consider user modeling. To overcome this challenge, \cite{tran2022ask} integrated reinforcement learning with dynamic user modeling to optimize the ordering of persuasive appeals, taking into account both the conversation history and the user's propensity to donate. Furthermore, numerous persuasive systems struggle with fluent communication and adaptability because of their heavy dependence on predefined strategies. An exception is~\cite{Furumai2024PersuaBot}, which extracts persuasive strategies from responses generated by large language models, identifies relevant supporting data, and produces factually accurate responses without compromising strategic effectiveness.

\subsection{Causal Inference in Persuasive Systems} 
Causal discovery methods~\cite{solus2021consistency, spirtes2013causal, spirtes2001causation} have evolved to address the complexities of understanding variable relationships. For instance, permutation-based techniques~\cite{solus2021consistency} reorder variables to test causal dependencies and identify the underlying causal model, such as the Ordering Search algorithm \cite{teyssier2012ordering} and Greedy Sparse Permutation (GSP) \cite{solus2021consistency}. These methods use permutations to explore causal structures, but often require strong assumptions and can be computationally intensive. To overcome these limitations, the Greedy Relaxations of the Sparsest Permutation Algorithm (GRaSP)~\cite{lam2022greedy} integrates a new permutation-based operation called \textit{tuck} that improves upon GSP by relaxing the  underlying true causal relationships without coincidental or hidden factors
and offering greater efficiency and accuracy in causal model search. GRaSP has proved particularly effective for dense causal graphs and large-scale models, outperforming many existing simulation algorithms~\cite{chickering2002optimal, ramsey2017million}. This makes GRaSP well-suited for datasets with many variables, such as our dataset with 27 and 23 variables related to the persuader and persuadee strategies, respectively. Identifying causal relationships between strategies is essential for optimizing persuasive dialogues and GRaSP's efficiency and accuracy make it a valuable tool for analyzing these relationships and enhancing persuasive dialogue systems. 

Counterfactual reasoning~\cite{hoch1985counterfactual} helps establish causality by comparing the observed outcome to a hypothetical scenario in which a key factor is changed. This enables better understanding of the impact of specific actions. As a core concept in causal inference, counterfactual reasoning helps quantify the effect of an intervention by imagining what would have happened in the absence of the intervention, which plays a critical role in establishing causal claims. In a persuasive system, this method can provide the reasoning process behind the system's responses that did not occur before~\cite{zeng2024counterfactual}.

In this work, we use counterfactual data for a reinforcement learning (RL) agent to possess a wide set of alternative actions to choose from and test them in order to learn the optimal policy to maximize the persuasion outcomes. The goal of RL~\cite{kaelbling1996reinforcement} in general is to train intelligent agents to make decisions in dynamic environments to maximize cumulative rewards over time. While several works have used RL effectively in persuasive systems (e.g.,~\cite{, Hiraoka2016, tran2022ask, Tiwari2022}), to our knowledge, we are the first to use RL together with causal reasoning in finding the optimal system dialog utterances (see our recent work in~\cite{zeng2024counterfactual} prior to this one). 

\section{Architecture}
\begin{figure}
    \centering
    \includegraphics[width=\linewidth]{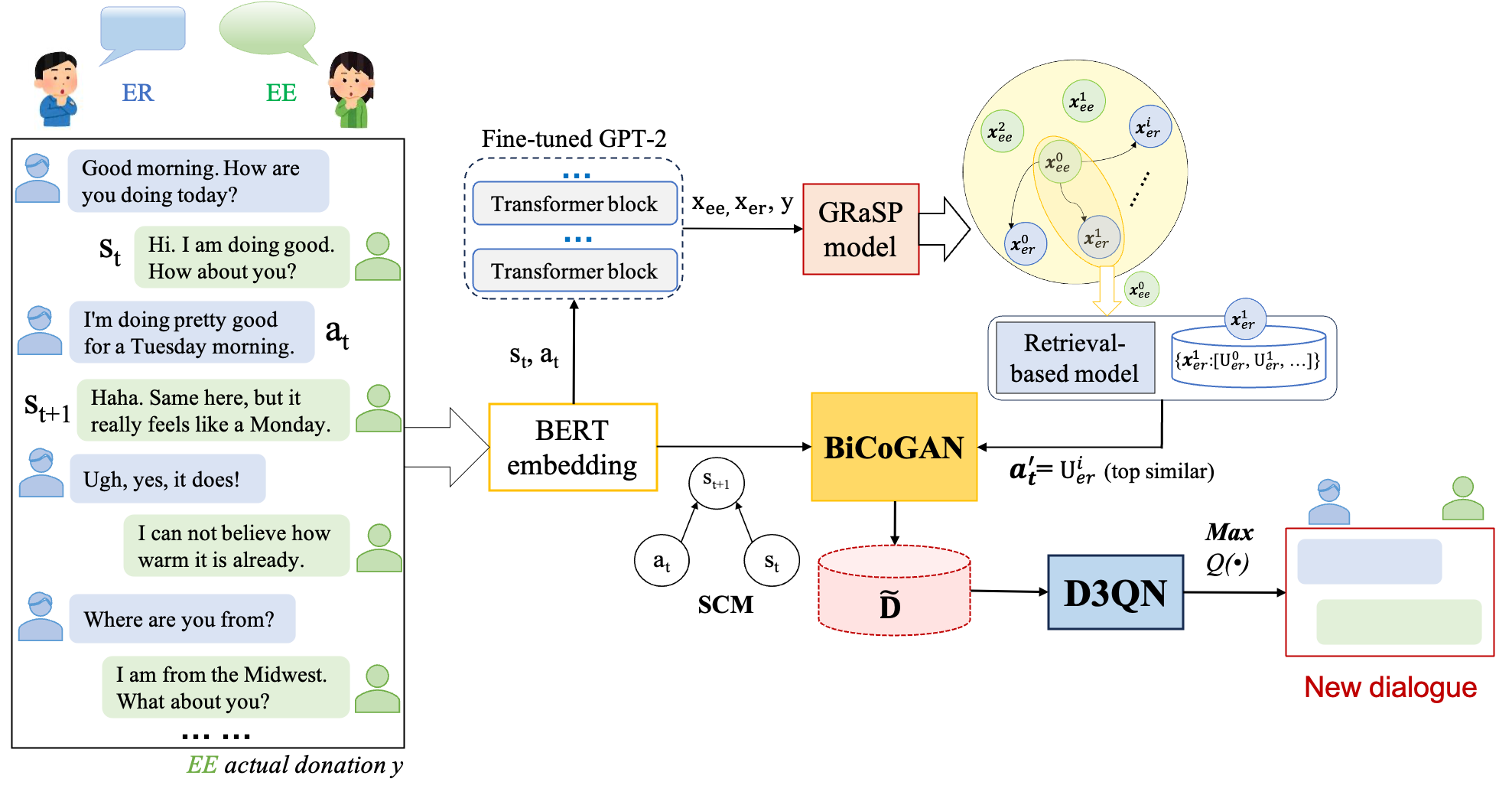}
    \caption{Illustration of our proposed architecture for optimizing persuasive dialogues. Each dialogue utterance is represented using BERT embeddings, while two fine-tuned GPT-2 models for persuadee and persuader predict the dialogue strategy associated with each utterance. The GRaSP method is employed for causal discovery to identify cause-effect relationships between the persuadee's (EE) and the persuader's (ER) strategies, which is influenced by the persuadee’s donation behavior. Specifically, we identify the causal relationship $x_{ee} \rightarrow x_{er}$, which is then incorporated into a retrieval-based model. This model selects the counterfactual action $a'_t$ by retrieving the most probable persuader utterance based on a similarity score with the current state $s_t$. To validate our hypothesis that persuasive dialogues adhere to an underlying structural causal mechanism, we utilize BiCoGAN to generate counterfactual data $\tilde{D}$. The counterfactual dataset is subsequently used to train a Deep Double Q-Network (D3QN), which learns an optimized policy aimed at maximizing Q-values. This process ultimately facilitates the generation of new dialogue strategies that have a higher likelihood of increasing donation amounts.}
    \label{fig:arch}
\end{figure}

We now introduce our problem setting, as well as the three components of our architecture (seen in Fig.~\ref{fig:arch}), namely, 
\begin{enumerate}
    \item causal discovery by GRaSP to identify the causal graph consisting of persuadee and persuader strategies that becomes the basis of counterfactual actions;
    \item generation of the counterfactual data using BiCoGAN that utilizes dialogue data along with the constructed counterfactual actions; and
    \item policy learning by D3QN to optimize persuasion policies and target outcomes.
\end{enumerate}

\subsection{Problem Setting}
Let $D = \{D_i\}$ be a recorded collection of dialogues, where each dialogue is represented as a sequence of transitions $D_i = \{(s_t, a_t, s_{t+1})\}_{t=0}^{T-1}$. Here, $s_t$ denotes the state at time $t$, $a_t$ is the action taken at time $t$, and $s_{t+1}$ is the resulting next state. This adheres to the edicts of a structural causal model (SCM)~\cite{pearl2000models}, a formalization for defining causal relationships between variables through structural equations. This enables us to obtain the causal relationships that help identify which actions will lead to the desired effects when implemented. 
We explore for Markov Decision Processes (MDPs), each characterized by $M = (S, A, f, R, T)$. Here, $S=\{s_{0}, s_{1}, s_{2}, ..., s_{\lceil T/2 \rceil}\}$ represents the finite set of states, while $A=\{a_{0}, a_{1}, a_{2}, ..., a_{\lfloor T/2 \rfloor}\}$ denotes the finite set of actions.
$R$ is the immediate reward, 
represented as numerical value received after acting in a specific state, essentially guiding the RL agent to make decisions that maximize performance with its environment.
The causal mechanism $f$ can now be formulated as
\begin{equation}
    s_{t+1} = f(s_{t}, a_{t}, \varepsilon_{t+1}), 
    \label{eq:SCM}
\end{equation}
\begin{wrapfigure}{r}{0.6\linewidth}
\includegraphics[width=\linewidth]{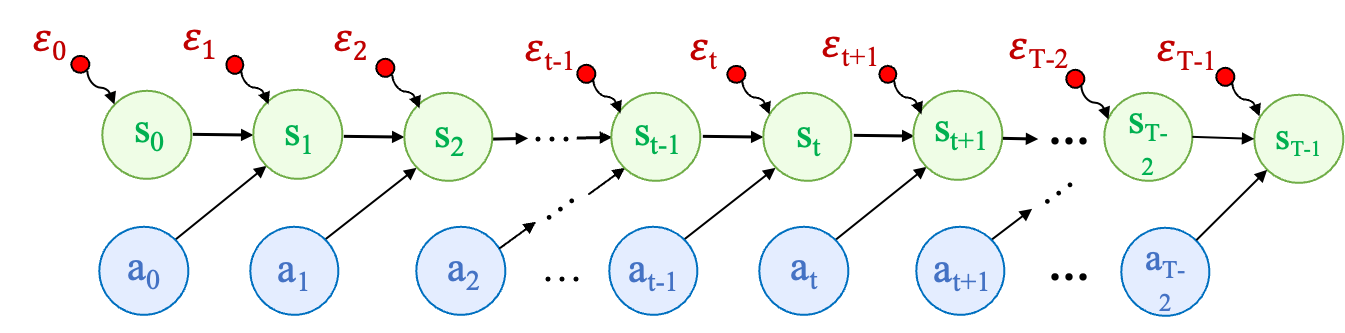}
\caption{Illustration of the state transition dynamics in a persuasive dialogue. The figure depicts how the persuadee’s state $s_t$ evolves based on the persuader’s action $a_t$, following a structured causal process. }
\label{transition}
\vspace{-10pt}
\end{wrapfigure}
\noindent where $s_{t}$ is the persuadee utterance, $a_{t}$ is the corresponding persuader utterance, both encoded as 768-dimensional BERT embeddings,
and $\varepsilon_{t+1}$ is the noise term independent of ($s_{t}$, $a_{t}$). We depict the transition process in Fig.~\ref{transition}. Lastly, given the real-world data $D$,
BiCoGAN generates 
$N$ counterfactual databases $\tilde{D}=\{\tilde{D}_{0}, \tilde{D}_{1}, \tilde{D}_{2}, ..., \tilde{D}_{N}\}$ in which $\tilde{D}_{i} =\{(s^{'}_{t}, a^{'}_{t})\}^{T-1}_{t=0}$. The aim is to determine the policy that would maximize the goodness of the resulting next states, as well as the persuasion outcomes, in $\tilde{D}$.
\subsection{Causal Discovery Using GRaSP} \label{causal_discovery}
Our approach utilizes the results from causal discovery to systematically inform the formulation of counterfactual actions, ensuring that the generated utterances are obtained in a reasonable way. We hypothesize there exists causal relationships between persuadee and persuader strategies that impact the flow of their dialogues and consequently the outcomes , i.e., the actual donations. By leveraging these causal relationships, we can adjust the strategies to maximize the final donation. 
We call on GRaSP~\cite{lam2022greedy} to identify these relationships. GRaSP has been found effective in handling a large number of variables, which makes it well-suited for our dataset. 

Our input data for GRaSP is comprised of the 27 persuader strategy variables $x_{er}$, 23 persuadee strategy variables $x_{ee}$, and the donation amount variable $y$. We use softmax probabilities to represent the 27-dim $x_{er}$ and 23-dim $x_{ee}$, and use as value for $y$ the actual donation amount that is min-max normalized. Our model then generates a vector of logits after applying the Transformer and classifier layers of the fine-tuned GPT-2 to the BERT-embedded persuader and persuadee utterances. Let $\mathbf{l}$ represent the logits for each option, i.e., $\mathbf{l} = [l_1, l_2, l_3, ..., l_i, ...]$, where $l_i$ is the logit score for option $i$. We convert the logits into probabilities by applying the softmax function. For a logit vector $\mathbf{l}$, we compute for $softmax(l_i) =\frac{e^{l_i}}{\sum_{j} e^{l_j}}$.
We use BERT embeddings to represent the utterances, and one-hot vectors to encode the corresponding strategies. 

In this work, we fine-tuned distinct GPT-2 models for the two roles, i.e., one each for the persuadee and persuader. These fine-tuned models are then used to generate the softmax probabilities for predicting the strategies. We employ GPT-2 to predict the strategies of the BERT-embedded utterances since it can effectively capture contextual relationships in dialogues, which makes it ideal for fine-tuning the persuadee and persuader strategies. It is known that GPT-2 has been trained on a vast dataset, which enables it to generate responses based on a wide range of topics without requiring extensive fine-tuning, making it also useful for general-purpose dialogue agents. It can also be fine-tuned on specific datasets to adapt to specific domains, as in the case of persuasive dialogues. However, since this part of our method can be ablated, other large language models can be explored based on their availability and cost.


\begin{figure}
    \centering
    \includegraphics[width=\linewidth]{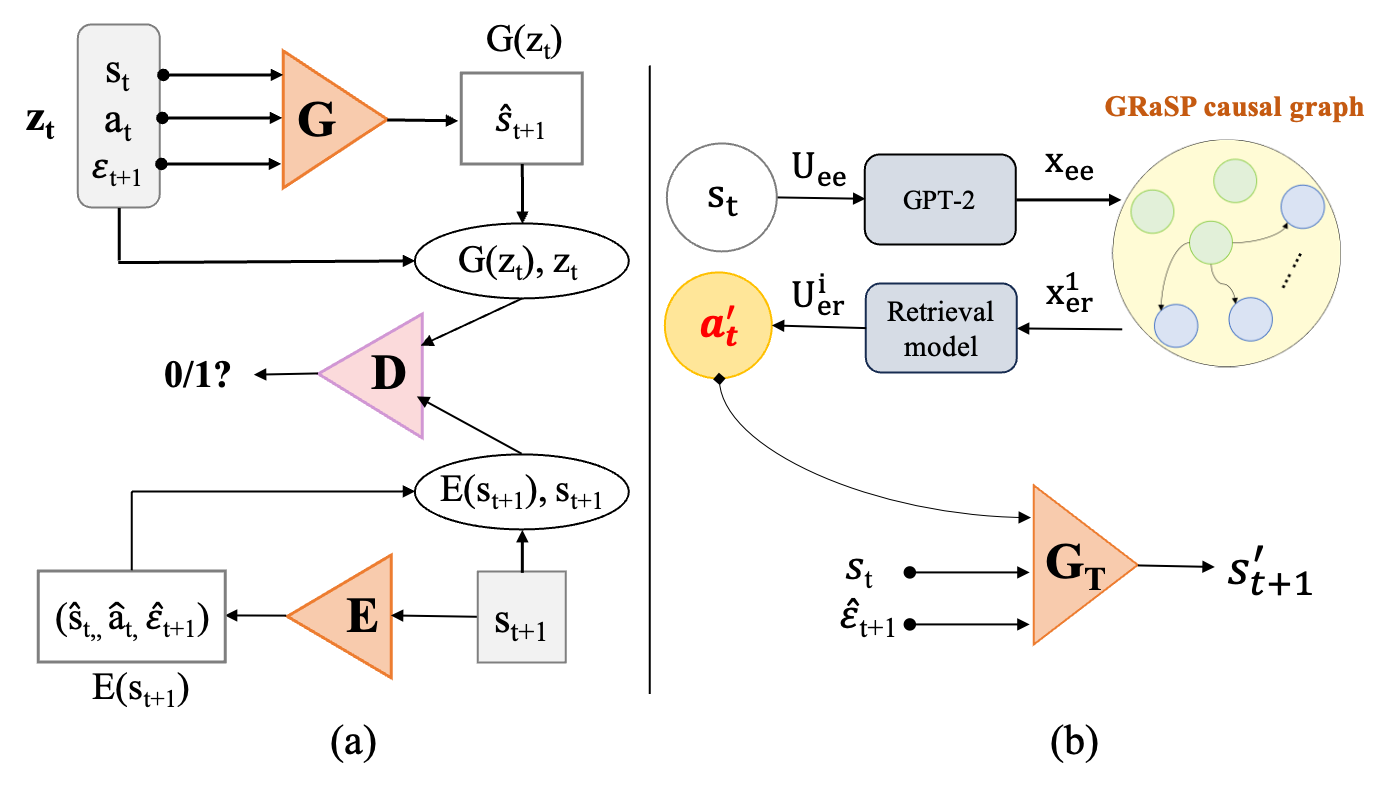}
    \caption{(a) The BiCoGAN training process, which consists of a generator ($G$), an encoder ($E$), and a discriminator ($D$). The generator learns to create realistic counterfactual samples by mapping latent representations to persuasive dialogue states, while the encoder reconstructs latent variables from generated samples, and the discriminator distinguishes between real and generated data.  
(b) Post-training, the fine-tuned GPT-2 model and a retrieval-based model are used for counterfactual action generation. A causal graph connects these models to determine the optimal alternative action $a^{'}_{t}$ for persuasion. The trained generator $G_T$ is then employed to construct the counterfactual next state $s^{'}_{t+1}$, enabling counterfactual reasoning for improved dialogue strategies.}

    \label{fig:causal_bicogan}
\end{figure}

Inspired by~\cite{tran2022ask}, we utilize two additional models to formulate the counterfactual actions that are connected utilizing the causal graph, namely, a fine-tuned GPT-2~\cite{radford2019language} and a retrieval-based model, shown in  Fig.~\ref{fig:causal_bicogan}(b). 
The connection between these two models is as follows.
The fine-tuned GPT-2 outputs the persuadee's strategy (the cause), which serves to help identify the corresponding persuader's strategy (the effect). The identified persuader's strategy is then used as input to the retrieval-based model component. We further explain these two models and their interaction.

The objective of the fine-tuned GPT-2 is to determine the strategy that the persuadee may be using (consciously or unconsciously) that led her to say what she did. We fine-tune the GPT-2 model to identify the persuadee's strategies using the 300 annotated dialogues of the PersuasionForGood dataset. Both the strategies and the associated utterances are included in these annotated dialogues. For example, the utterance ``Save the Children [the charitable organization] has not convinced me they can work towards accomplishing those tasks'' is an instance of the strategy \textit{negative-reaction-to-donation} (see Fig.~\ref{fig:example_causal} for more examples of the strategy-utterance association). We then use the fine-tuned GPT-2 model to predict the strategies of the remaining unannotated 717 dialogues of PersuasionForGood. 
Using the 300 annotated dialogues, we performed five-fold cross-validation to verify the model's efficacy, achieving an accuracy of 92.3\%. The 300 annotated dialogues and the remaining 717 unannotated ones were compared for intra- and inter-strategy similarity. The measured similarities for the annotated set are 0.632 and 0.559, respectively; while the results for the unannotated set are 0.645 and 0.576, respectively. Hence, both sets share a lot of similarities both within and between strategies. These results verify the GPT-2 model can strongly predict the strategies.

The retrieval-based model, on the other hand, aims to obtain the persuader utterances in natural language by using the pertinent persuader strategies as a query from the causal graph. Once the persuader's strategy is identified from the causal graph, the retrieval-based model then converts this strategy into natural language by utilizing the TF-IDF weighted bag-of-words vector to represent the utterance, as seen in the Fig.~\ref{fig:causal_bicogan} (b), from persuader's strategy $x^1_{er}$ leading to $U^i_{er}$. Given the cause-effect pairs ${x_{ee}} \rightarrow {x_{er}^{i}}$, where $i=1,2, ..n$, $n$ is the number of effects impacted by the common cause $x_{ee}$, we first randomly select one effect from a pair, then, use the cosine similarity score to select an appropriate persuader response. 
Given state $s_t$, We rank all prospect persuader utterances from $\{U_{er}\}$, each associated with effect strategy $x_{er}$, based on the similarity between each $U_{er}$ and persuadee utterance $U_{ee}$ with cause strategy $x_{ee}$.
The counterfactual action $a^{'}_{t}$ is then determined by selecting at random one of the top three utterances with the highest similarity scores. By randomizing, we prevent choosing over and over again the same topmost utterance.


\subsection{Generation of the Counterfactual Dataset~$\tilde{D}$} \label{counter_data}

We assume state $s_{t+1}$ adheres to the SCM in Equation (\ref{eq:SCM}). 
We then apply BiCoGAN (Fig.~\ref{fig:causal_bicogan}) to derive function $f$ by reducing the disparity between the input \textit{actual} data and the generated counterfactual data, ensuring that the counterfactual states remain realistic and plausible given the observed, factual scenarios.
At the same time, it also estimates the value of the noise term $\varepsilon_{t+1}$ (seen in Fig.~\ref{fig:causal_bicogan}), which represents the disturbances arising from unobserved factors. 
Specifically, BiCoGAN learns from two directions: (1) estimating the next state, $\hat{s}_{t+1}$, from ($s_{t}$, $a_{t}$, $\varepsilon_{t+1}$) via $G$, and (2) deriving ($\hat{s}_{t}$, $\hat{a}_{t}$, $\hat{\varepsilon}_{t+1}$) from $s_{t+1}$ via $E$. The discriminator $D$ discerns between real and generated data.
The generator and encoder each have distinct formulations for their respective distributions:
\begin{equation}
\begin{aligned}
    P(\hat{s}_{t+1}, s_{t}, a_{t}, \varepsilon_{t+1}) &= P(s_{t}, a_{t}, \varepsilon_{t+1})P(\hat{s}_{t+1}|s_{t}, a_{t}, \varepsilon_{t+1}), \\
    P(s_{t+1}, \hat{s}_{t}, \hat{a}_{t}, \hat{\varepsilon}_{t+1}) &= P(s_{t+1})P(\hat{s}_{t}, \hat{a}_{t}, \hat{\varepsilon}_{t+1}|s_{t+1}),
\end{aligned}
\end{equation}
where $\hat{s}_{t}$, $\hat{s}_{t+1}$, $\hat{a}_{t}$, and $\hat{\varepsilon}_{t+1}$ are estimations of $s_{t}$, $s_{t+1}$, $a_{t}$, and $\varepsilon_{t+1}$. Deceiving the discriminator requires optimizing the loss function as a minimax game:
\begin{equation}
\begin{aligned}
\min_G \max_D V(D, G, E) &= \min_G \max_D \{\mathbb{E}_{s_{t+1} \sim p_{\text{data}}(s_{t+1})}[\log D(E(s_{t+1}), s_{t+1})] \\ &+ \mathbb{E}_{z_{t} \sim p(z_{t})}[\log(1 - D(G(z_{t}), z_{t}))] \\& +\lambda E_{(s_{t}, a_{t}, s_{t+1}) \sim p_{\text{data}}(s_{t}, a_{t}, s_{t+1})}  [R((s_{t}, a_{t}), E(s_{t+1}))]\},
\end{aligned}
\end{equation}
where $z_{t} = (s_{t}, a_{t}, \varepsilon_{t+1})$, and $R$ regularizes via its hyper-parameter $\lambda$.

After learning the SCM, the trained BiCoGAN now embodies the understanding of the dialogue process.
In particular, new counterfactual dialogues are generated by feeding counterfactual actions into the trained generator $G_{T}$.
The trained GPT-2 model is used to forecast the persuadee's strategy from a given state, which is the first step in creating the counterfactual actions that would make up the persuader's utterances. 
Afterwards, the causal graph is explored to find every possible persuader strategy linked to this predicted persuadee strategy. The counterfactual action is chosen from among the persuader strategies that have the highest similarity score given the context (refer to Section~\ref{causal_discovery}).

Given the tuple ($s_{t}$, $a_{t}$, $s_{t+1}$) at time $t$, the aim is to determine the alternative next state $s^{'}_{t+1}$ when a counterfactual action $a^{'}_{t}$ is instead performed. If the strategy in $s_{t}$ as cause is followed by a corresponding effect strategy, then we use the above methods to create the counterfactual actions. Otherwise, i.e., no cause and effect pair exists, we randomly select $a^{'}_{t}$ from the persuader's utterances. Once we have obtained  $a'_{t}$, we use ($s_{t}$, $a^{'}_{t}$) as input to $G_{T}$, which then outputs $s^{'}_{t+1}$ that is used to form the counterfactual database $\tilde{D}$ (depicted in Figs.~\ref{fig:causal_bicogan}(b) and~\ref{fig:causal_bicogan}(c)). We formulate $\tilde{D}$ as follows:
\begin{equation}
\begin{aligned}
\tilde{D} &= \{\tilde{D}_{0}, \tilde{D}_{1}, ..., \tilde{D}_{i}, ..., \tilde{D}_{N-1}\} \\
&= \{(s^{'}_{0}, a^{'}_{0}, s^{'}_{1}, a^{'}_{1}, ..., s^{'}_{t}, a^{'}_{t}, ..., s^{'}_{T-1})^{M-1}_{j=0}\}^{N-1}_{i=0},
\end{aligned}
\end{equation}
where $T$ denotes the total number of time steps, $M$ indicates the number of dialogues, and $N$ represents the number of counterfactual databases.


\subsection{Policy learning}\label{policy}
\begin{figure}[t!]
\centering
\includegraphics[width=\textwidth]{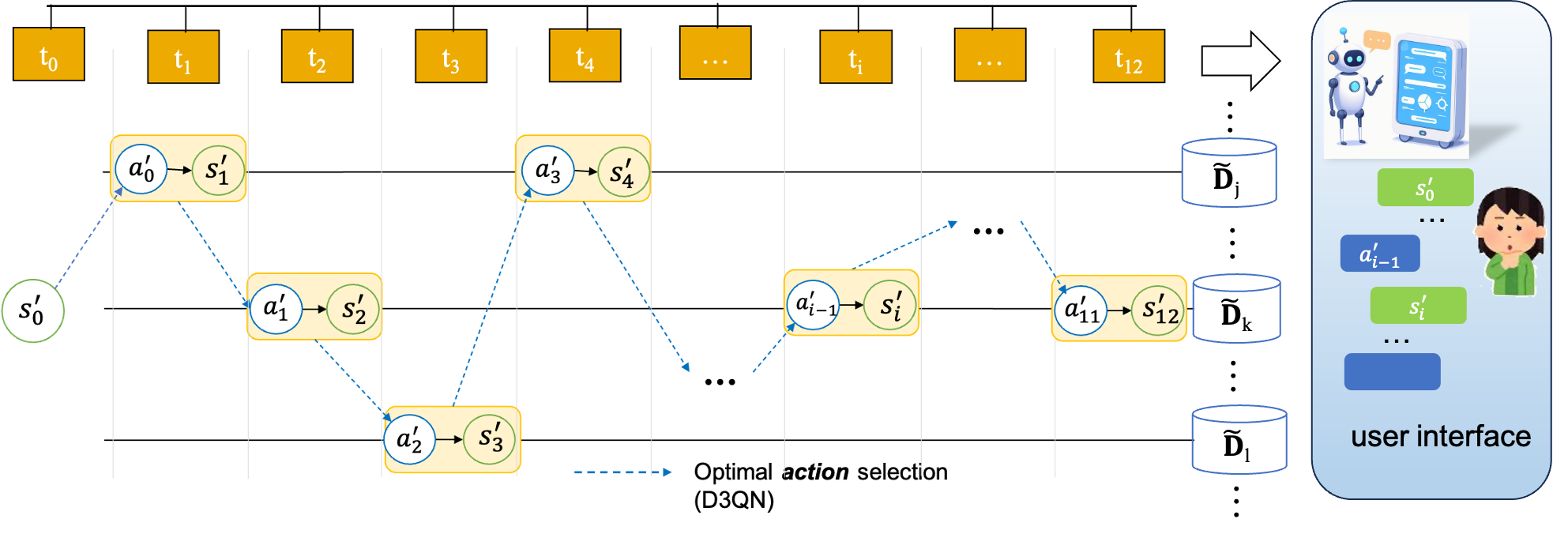}
\caption{The process of generating counterfactual dialogue sequences to improve reward prediction in policy learning. Starting from the initial state $s_0^{'}$, a sequence of action-state pairs ($a^{'}_{i-1}$, $s^{'}_i$) is generated. For each state $s_t^{'}$, the next action $a^{'}_i$ is selected by maximizing the Q-value: $k = \arg \max(Q(s_t^{'}, a_t^{'}))$ for $t = [0..N]$, where $N = 1$. Actions are drawn from the $t$-th dialogue in the counterfactual dataset $\tilde{D}_k$. The dialogue begins with the first several utterances from the ground truth dialogue and continues until the maximum length of 25 utterances is reached.}
\label{fig:newdialogue}
\end{figure}

After generating the counterfactual dataset $\tilde{D}$, Dueling Double Deep Q-Network (D3QN)~\cite{raghu2017deep} learns the policies on $\tilde{D}$ that maximize the expected future rewards. D3QN is an improved method of Deep Q-Networks (DQNs)~\cite{mnih2015human}, designed to address the issue of Q-value overestimation. To achieve this, D3QN decomposes the Q-value into a state-dependent advantage function $A^{\pi}(s, a)$, which quantifies how much better a specific action is compared to alternatives, and a value function $V^{\pi}(s)$ representing the expected reward from state $s$. The Q-value over $\tilde{D}$ is then computed as:
\begin{equation}
    \begin{aligned}
    Q(s', a';\theta) = \mathbb{E}\left[r(s', a') + \gamma \max_{a'} Q(s', a';\bar{\theta}) \mid s^{'}, a^{'}\right],
    \end{aligned}
    \label{qlearning}
\end{equation}

\noindent where $r(s^{'}, a^{'})$ is the reward of taking counterfactual action $a^{'}$ at counterfactual state $s^{'}$, $\gamma$ is the discount factor of the maximum Q-value among all possible actions from the next state, and $\bar{\theta}$ and $\theta$ are the weights of the target network and main network, respectively. The reward function $r(s^{'}, a^{'})$
that calculates the immediate reward value is an LSTM-based architecture trained using the dialogues and the donation values coming out of these dialogues. The reward value for ($s'_{t}$, $a'_{t}$) is calculated by our Dialogue-based Donation Prediction (DDP) model as follows:
\begin{equation}
  r(s', a')=\left\{
  \begin{array}{@{}ll@{}}
    0, & \text{if}\ t<T-1, \\
    DDP(d) = LSTM(BERT_{embedding}(\{s'_{t}, a'_{t}\}^{T-1}_{t=0})), & \text{otherwise},
  \end{array}\right.
  \label{reward_fun}
\end{equation} 
where $d$ denotes the counterfactual state-action transitions as follows:
\begin{equation}
    d=(s^{'}_{0}, a^{'}_{0}, s^{'}_{1}, a^{'}_{1}, ..., s^{'}_{t}, a^{'}_{t}, ..., s^{'}_{T-1}).
\end{equation}
The target Q-values are obtained from actions through a feed-forward pass on the main network, rather than being directly estimated from the target network. During policy learning, the state transition begins with state $s^{'}_{0}$. The optimal counterfactual action $a^{*}_{0}$ corresponding to the state $s'_{0}$ is identified by computing for
\begin{equation}
    a^{*}_{0} = \arg \max_{a'_{0i}} Q(s'_{0}, a'_{0i}; \theta, \alpha, \beta),
\end{equation}
where $i = 0, 1, 2, \ldots, N-1$, and $N$ represents the number of counterfactual databases. This action selection process is illustrated in Fig.~\ref{fig:newdialogue}.



We hypothesize that incorporating counterfactual data will enhance the donation amounts during policy learning with D3QN. For this, we use Mean Squared Error (MSE) as the loss function to update the weights of D3QN, optimizing this function once for each dialogue.
The subsequent steps involve training policies on $\Tilde{D}$ to optimize the Q-values and improve the predicted future cumulative rewards over the ground truth. 

We use the same method as Equation (\ref{reward_fun}) to predict the reward (donation amount) for each dialogue. Specifically, we compute the cumulative sum of the predictions from the trained reward model as follows:
\begin{equation}
\begin{aligned}
  \hat{R}_{i} &= DDP(d) = LSTM(BERT_{embedding}(\{s'_{0}, a'_{0}, s'_{1}, a'_{1}, ..., s'_{T-1}\})), \\
  R_c &= \left[ \sum_{i=0}^{0} \hat{R}_i, \sum_{i=0}^{1} \hat{R}_i, \ldots, \sum_{i=0}^{k-1} \hat{R}_i, \ldots, \sum_{i=0}^{M-1} \hat{R}_i \right],
  \label{creward_fun}
\end{aligned}
\end{equation} 
where $M$ is the total number of dialogues, \(R_c\) is the list of cumulative rewards over $M$ dialogues, and each element $R_c^k = \sum_{i=0}^{k-1} \hat{R}_i$ indicates the cumulative reward over \(k\) dialogues. 


\section{Experiment} \label{experiment}
The  $PersuasionForGood$~\footnote{https://convokit.cornell.edu/documentation/persuasionforgood.html} dataset of interpersonal dialogues was built around the idea of influencing user behavior.
Specifically, in the online persuasion tasks, the $persuader$ (ER) was required to influence the $persuadee$ (EE) to donate to an actual charitable organization called Save the Children~\footnote{https://www.savethechildren.org/}. 
A total of 1,017 dialogues comprise the dataset, 300 of these were annotated with 50 different strategies in the categories of \textit{general conversation behavior} (23 strategies) for EE, and \textit{appeal} (7), \textit{inquiry} (3), and \textit{task-related non-persuasive} (17) for ER.
While each dialogue also contains the donations of both EE and ER, our primary focus in this work is EE's donation behavior as it directly relates to the persuasion outcome. 
To ensure the robustness of the reward model, we deliberately excluded dialogues with donation amounts exceeding \$20.00. This decision aims to reduce potential bias and promote more equitable model training. As a result, 997 dialogues remained, each containing 25 exchanges alternating between EE and ER.

In our experiments, we apply BERT embeddings derived from a pre-trained BERT model~\cite{devlin2018bert} available on Hugging Face\footnote{https://huggingface.co/bert-base-uncased}. A 768-dimensional BERT feature vector is employed to represent each natural language utterance made by either EE or ER. Additionally, our approach consists of three models that are trained independently with different input data. We set the dimension of the noise term in BiCoGAN to match the 768-dimensional BERT embeddings. As a result, we generated a total of 100 counterfactual dialogues using various sets of hypothetical actions.

\begin{figure}[p]
    \centering
        \centering
        \includegraphics[width=\linewidth]{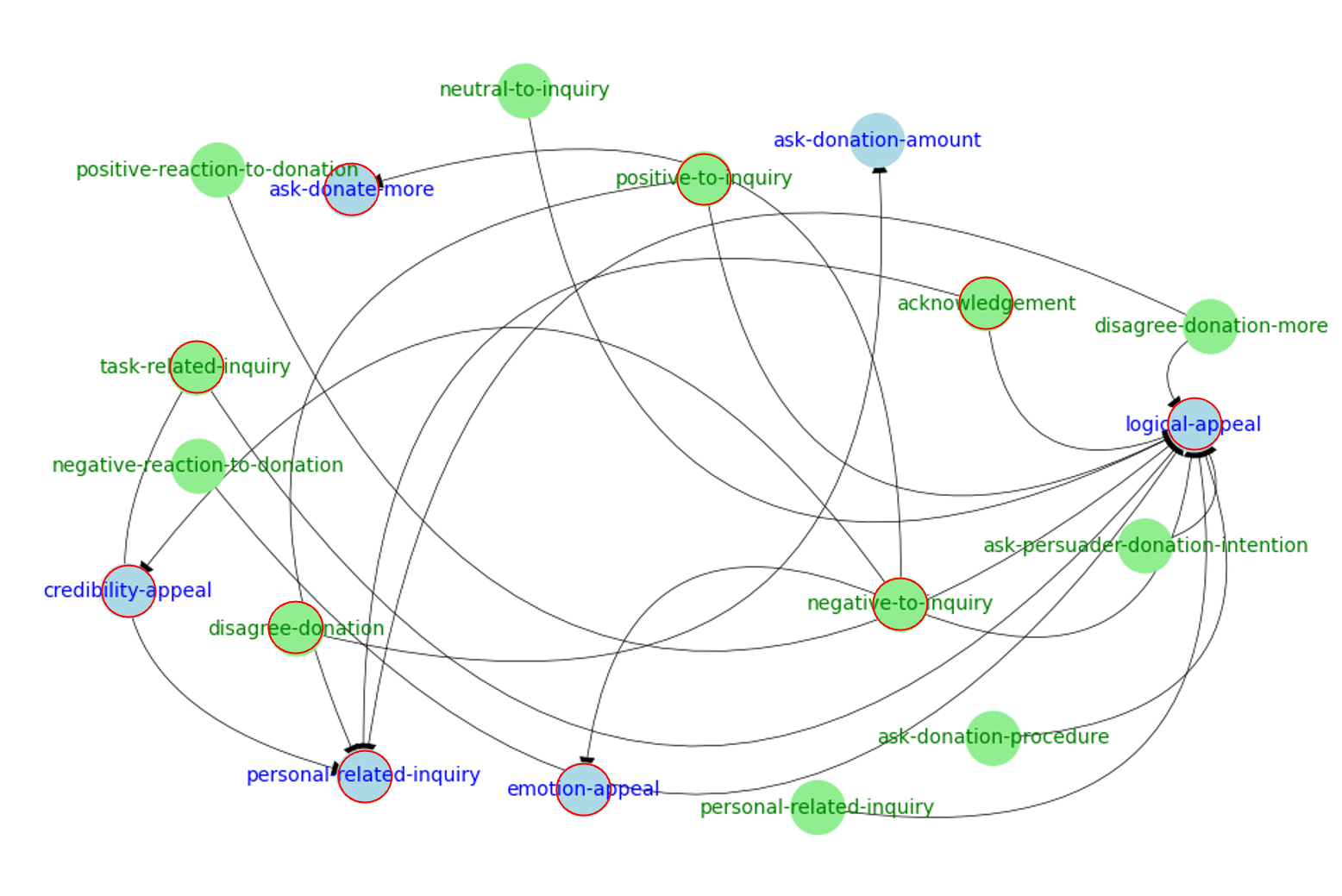}
        \label{fig:second_subfigure}
    \caption{This causal subgraph is obtained using GRaSP, which illustrates the causal relationships between strategies. The arrows pointing from EE to ER indicate that the EE strategies cause the ER strategies and vice versa. EE strategies, highlighted with a green background circle, act as causes in the dialogues, influencing the ER strategies.  ER strategies, shown with a blue background, represent the outcomes shaped by EE's causal effects. Circles with red outlines are the strategies discussed in this paper.}
    \label{fig:causal_graph}
\end{figure}


\section{Results}
\label{results}

To generate the counterfactual database $\Tilde{D}$, we submit to BiCoGAN the counterfactual action set~\{$a^{'}_{t}$\}, which we select using two different approaches: (1) randomly choose from ER's utterances or (2) select per the discovered causal relations. For (1), our initial step involves constructing \{$a^{'}_{t}$\} by randomly selecting in PersuasionForGood the actual action set \{$a_{t}$\}. 
On the other hand, for (2), we use the causal discovery methods to systematically identify the cause-effect relationship pairs plausibly underlying the strategies of EE and ER. In effect, with (1), we implement a baseline that provides a simple and effective starting point for comparison, relying on traditional methods for generating counterfactual data without incorporating causal knowledge. This makes it a clear reference for evaluating the added value of our causality-driven approach. Derived from~\cite{zeng2024counterfactual}, this baseline shares key elements with our generative framework, including components, tasks, and workflow - to ensure a fair comparison. While the baseline method enhances persuasive actions by generating counterfactual data, it does not leverage causal discovery. In contrast, our method uses causal insights to improve the quality of counterfactual data and optimize alternative actions, ultimately improving the persuasion outcomes. By targeting strategy-level causal links, we can pinpoint the causal effects of EE strategies on the ER responses. For example, when EE expresses \textit{disagree-donation} or \textit{negative-to-inquiry}, which is indicative of the hesitancy to donate, by knowing which ER response is most likely elicited by this would allow testing first and foremost the effectiveness of this ER response. Thus, D3QN does not need to test every possible ER response arbitrarily, which would require considerable time before converging on an optimal response policy. 

In observing the causal graph from GRaSP, we found that \textit{logical appeal} emerges as a prominent strategy across 18 pairs, predominantly used by ER in response to various EE tactics, including \textit{positive-to-inquiry} and \textit{task-related-inquiry}. Our findings reveal that \textit{logical appeal} co-occurs with \textit{positive-to-inquiry} in 238 dialogues and with \textit{task-related-inquiry} in 271 dialogues, highlighting its vital role in satisfying the information needs of the EE. Furthermore, the \textit{personal-related-inquiry} strategy is often adopted by ER following EE's \textit{positive-to-inquiry} and \textit{acknowledgement} strategies. Our analysis shows that 263 dialogues reflect the overlap between \textit{personal-related-inquiry} and \textit{positive-to-inquiry}, while 172 dialogues exhibit co-occurrences with \textit{acknowledgement}, emphasizing its significance in tailoring dialogues to individual needs. Additionally, EE's \textit{closing} strategies prompt ER to utilize \textit{credibility-appeal}, as demonstrated by 265 dialogues that show these strategies working together, indicating their effectiveness in reinforcing key messages and achieving persuasion at the conclusion of discussions. Lastly, ER frequently employs \textit{emotion-appeal} in reaction to EE's \textit{negative-to-inquiry} strategies, with 265 dialogues revealing the interplay between \textit{emotion-appeal} and \textit{credibility-appeal}, aimed at fostering an emotional connection when logical arguments are inadequate. It is also important to note that while the donation amount is included among GRaSP's inputs, our primary interest lies in the causal relationships among the strategies. Determining whether the strategies employed by ER will lead to the desired donations is the responsibility of D3QN, which will evaluate the effectiveness of these strategies based on their impact on cumulative rewards and Q-values.

\begin{figure}
    \centering
    \includegraphics[width=\linewidth]{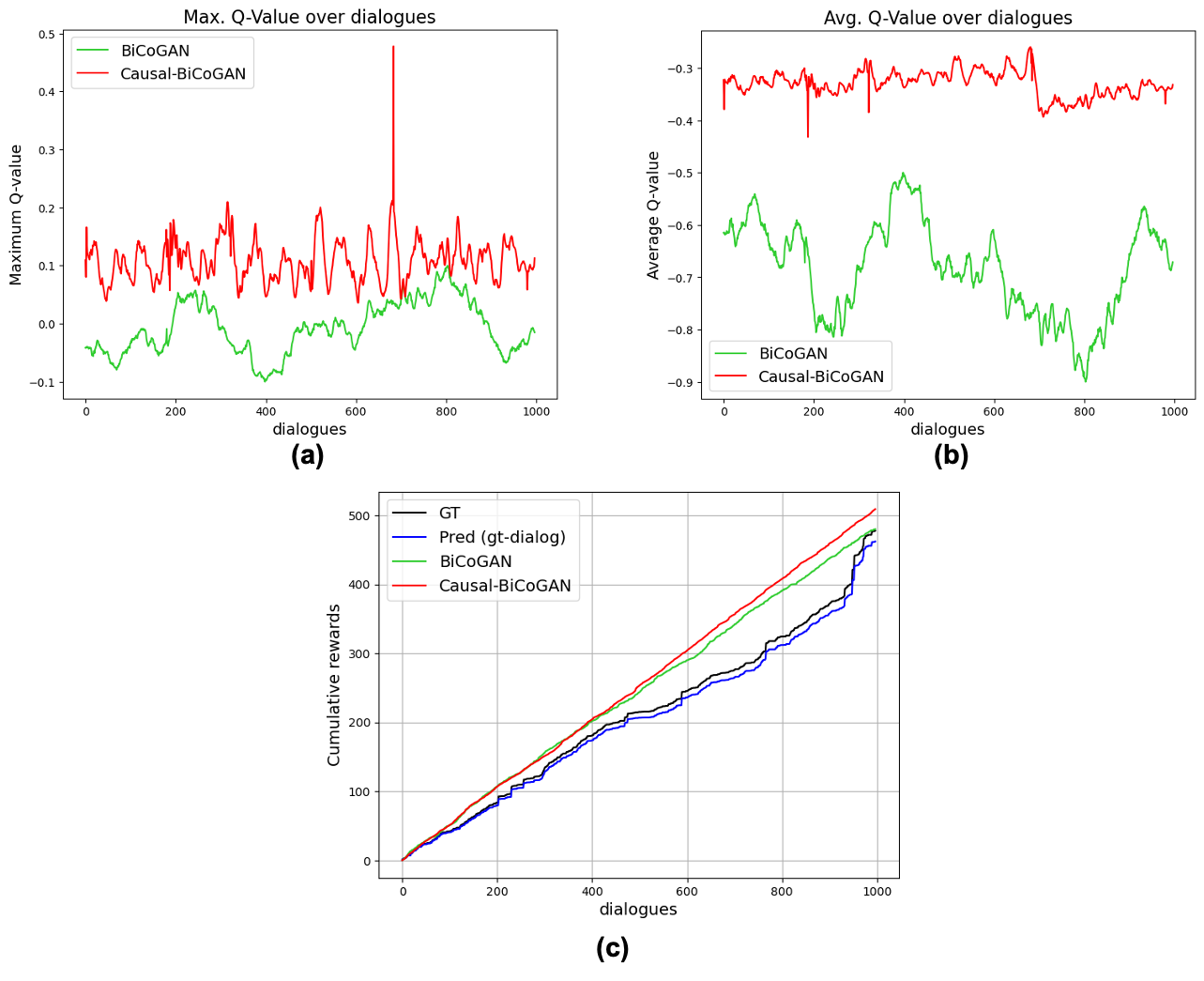}
    \caption{(a) and (b). The \textit{maximum} and \textit{average} Q-values, respectively, in dialogues with counterfactual cases, comparing two methods. In (a), the maximum Q-values are computed for dialogues where counterfactual cases are generated using the BiCoGAN model (legend: BiCoGAN) and the GRaSP-based BiCoGAN method (legend: GRaSP+BiCoGAN). Similarly, in (b), the average Q-values are computed for the same two methods: BiCoGAN and GRaSP+BiCoGAN, which support the creation of the counterfactual cases. (c) The predicted donation amount, $DDP(D^*)$, after applying the D3QN model to the counterfactual data, $\tilde{D}$, demonstrating the effect of counterfactual reasoning on donation predictions.}
    \label{fig:three_RLresults}
\end{figure}

Following post-policy learning by D3QN, we obtain optimized counterfactual dialogues that serve as plausible alternatives to the real-world dialogues in the dataset. 
Specifically, the sequences of real-world dialogues commence at $s_{0}$, whereas the alternative-world dialogues begin at $s^{'}_{0}$, with the condition that $s_{0}$ = $s^{'}_{0}$. During the evaluation of the Q-values associated with the learned optimal policy, the approach utilizing BiCoGAN with causal discovery yielded significantly higher Q-value estimates, as demonstrated in both the maximum (Fig.~\ref{fig:three_RLresults}a) and average (Fig.~\ref{fig:three_RLresults}b) estimates, compared to BiCoGAN not utilizing the discovered causal relations. 
We also show the cumulative predicted rewards $DDP(D*)$ (Fig.~\ref{fig:three_RLresults}c), $D*$ being the input dialogues. We see that \textbf{GRaSP+BiCoGAN} achieves the best results overall, and both \textbf{BiCoGAN} and \textbf{GRaSP+BiCoGAN} have higher cumulative donation amounts than the ground truth. Specifically, the total predicted donation amount is \$508.78 (an increase of \$31.16) for \textbf{Causal+BiCoGAN}. For \textbf{BiCoGAN}, the predicted amount is \$496.15 (an increase of \$18.53). These amounts are higher than the ground truth actual donation amount of \$477.62. These numbers indicate that identifying strategy-level causal relations between EE and ER can generate counterfactual strategies that improve persuasion policies and persuasive dialogue outcomes. It is also important to note that while D3QN effectively mitigates the issues associated with unusually high Q-values~\cite{raghu2017deep}, we observe in Fig.~\ref{fig:three_RLresults} (a) and (b) that one dialogue exhibits a significantly higher maximum Q-value than the others, while the mean Q-value remains comparable to the rest. One potential factor contributing to this outlier is the model's ability to identify a strong association between the selected response utterance for ER and relatively high outcomes. This phenomenon likely arises from sufficient exploration of this particular dialogue, allowing the model to focus on high-reward responses.

\begin{figure}
\centering
\vspace{-15pt}
\includegraphics[width=0.9\textwidth]{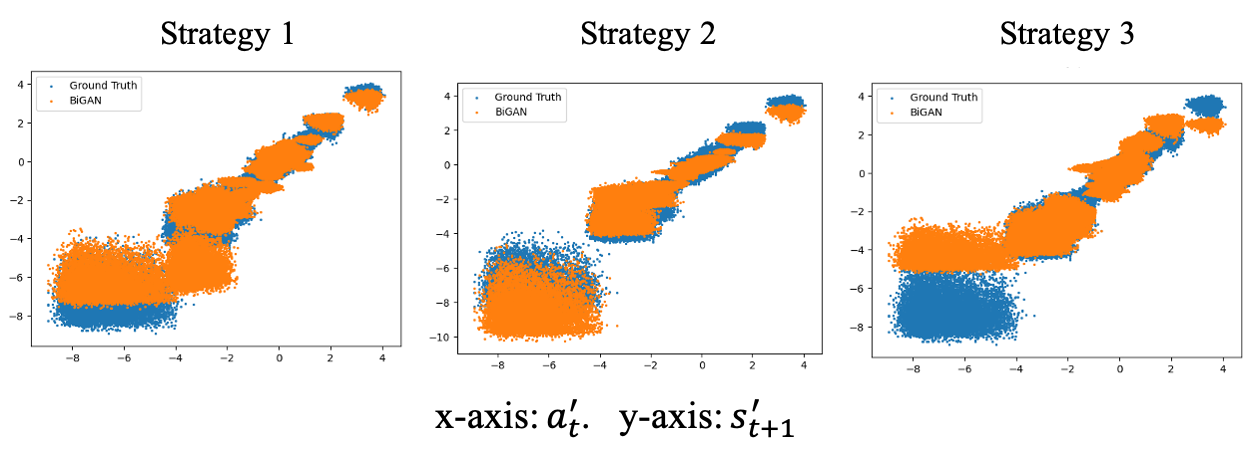}
\caption{The relationship between the counterfactual action $a^{'}_{t}$ and the next state: The counterfactual case $s^{'}_{t+1}$ (in orange) is generated by BiCoGAN, and the ground truth next state $s_{t+1}$ (in blue) represents the actual outcome. We observe that the counterfactual state $s^{'}_{t+1}$ closely aligns with the real state $s_{t+1}$ when Strategy 2 is applied. This alignment indicates that the generated alternative world scenarios, although different from the actual dialogue sequence, are not significantly different from reality. In other words, the counterfactual scenarios remain realistic and applicable to the real world, demonstrating that the causal inference from the model provides plausible outcomes that are grounded in real-world interactions.}
\label{fig:3bicogan}
\vspace{-15pt}
\end{figure}

To validate the different cases of counterfactual action selection, we compare the impact of selecting the utterances randomly and guiding the selection using the information provided by the resulting causal graph. Furthermore, we also discovered from our experiments that 
the position of the \textit{greeting} utterances of ER have consequences on the quality of counterfactual data. It seemed a trivial matter to us at the start, but our results proved us wrong (later to this in Fig.~\ref{fig:3bicogan}).
Since the great majority of dialogues in PersuasionForGood start with ER greeting EE, 
we also account for this in our three strategies for creating the alternative-world action set \{$a^{'}_{t}$\} based on the real-world action set \{$a_{t}$\} that represent ER's utterances, as follows:


\begin{itemize}
    \item Strategy 1: We sample from all the utterances in $\{a_{t}\}$ into the set 
 $\{a^{'}_{t}\}$. 
    \item Strategy 2: We remove the initial utterance of ER $\{a_{0}$\} in each dialogue, and then sample from the remaining actions in $[\{a_{t}$\} - $\{a_{0}$\}] into $\{a^{'}_{t}\}$.
    \item Strategy 3: We eliminate the first three ER utterances $\{a_{0}, a_{1}, a_{2}$\} from each dialogue,
    and sample from the rest of actions in $[\{a_{t}$\} - $\{a_{0}, a_{1} a_{2}$\}] into $\{a^{'}_{t}\}$.
\end{itemize} 

In the context of Strategy 1 in which all greeting utterances are included, random sampling would position the greetings anywhere - in the beginning, middle, or end - of the counterfactual dialogues. In the case of Strategy 2, the greetings are not included when sampling, thereby preserving the occurrence of the greetings at the beginning of the dialogues. We use the same reasoning in Strategy 3, but also skipped two additional initial utterances as some dialogues may include extended greetings at the start.

To assess the effectiveness of the counterfactual data generated by the three strategies mentioned above, we compare the counterfactual states produced by BiCoGAN with the corresponding ground truth states (Fig.~\ref{fig:3bicogan}). We can see that the counterfactual data from Strategy 2 has more data points that are closely aligned with, but not the same as, the actual scenarios, contrast to the counterfactual data from the other two strategies. This means that the alternative scenarios constructed via Strategy 2 are more realistic alternatives. Thus, we are correct to position the greetings only at the beginning of the dialogues. It is interesting that this simple assumption could lead to more plausible alternative-world scenarios. On the other hand, the less satisfactory results from Strategy 1 are understandable, as greetings that appear in the middle or at the end of conversations - when they should have been at the beginning - seem odd. It also conflicts with the social norm of greeting at the start of conversations, especially when introducing oneself to a stranger if only to solicit donations. With Strategy 3, we may have shuffled to somewhere else in the counterfactual dialogues important utterances that are not greetings but have to be at the beginning of the conversations, which consequently led to less realistic alternative scenarios. 
These suggest that maintaining the traditional conversational norm of greeting at the start enhances the effectiveness of counterfactual reasoning.

\section{Discussion}\label{discussion}

Constructing causal models involves uncovering the underlying physical processes responsible for generating the observed data and encoding them within the model~\cite{Pearl2009}. Consequently, it should be possible to measure the causal impact of a specific action. In particular, when an intentional action is taken to alter a given situation, the model should generate counterfactuals - hypothetical scenarios that represent alternative realities. A key factor in achieving success here is that the derived counterfactual data is augmented with a substantial amount of unobserved interaction data, addressing the limitations of a fixed environment that may not accurately reflect what happens in the real world. This approach not only helps mitigate data scarcity but also introduces heterogeneity across users and situations~\cite{lu2020sample}. Regarding whether our method generates a plausible alternative world through these hypothetical scenarios, our results from Strategy 2 in Fig.~\ref{fig:3bicogan} suggest that the alternative reality data points do not diverge significantly and remain identifiable from the observed real-world conditions.

By positioning the causal discovery method as a source of causally related elements that serve as inputs to counterfactual reasoning, we address the challenge of characterizing heterogeneity through structured augmentation, rather than generating hypothetical scenarios ad hoc. In other words, causal discovery facilitates counterfactual reasoning that is less biased and grounded in causal evidence, thereby enhancing the plausible validity of conclusions drawn from hypothetical scenarios. We measured this by comparing the results with and without causal discovery guiding counterfactual reasoning. As shown in our results in Fig.~\ref{fig:three_RLresults}, the former indeed improved both the persuasive policies and the outcomes of persuasion.

Our findings demonstrate that incorporating causal discovery provides more accurate counterfactual scenarios, leading to dynamic persuasive strategies. Unlike traditional systems that rely on static strategies, our model dynamically adapts to user states. GRaSP enhances our understanding of persuasion-response relationships, while BiCoGAN generates plausibly realistic counterfactual utterances, resulting in higher cumulative rewards and Q-values. Compared to prior works, such as~\cite{zeng2024counterfactual}, which uses BiCoGAN for counterfactual generation and reinforcement learning for policy optimization, our approach uniquely embeds causal discovery. This allows us to capture user variability as we account for causal interactions effectively, enabling greater adaptation and improved persuasion outcomes.

We believe that our work has broader implications for its applicability. This research demonstrates that persuasive systems can move beyond static models and predefined strategies typically tested in controlled environments. By incorporating causal reasoning and dynamically adapting to user states, these systems can offer more flexible, context-aware, and personalized interactions. This approach has significant implications for applications in marketing, health communication, and social good initiatives (e.g., in~\cite{Furumai2024PersuaBot}). For instance, healthcare data often contains limited records for each patient, which hampers further analysis, and patients may respond differently to the same known treatments. These factors can hinder the optimization of medical intervention policies. Moreover, our method allows for innovation in intervention strategies using augmented data alone - eliminating unethical practices, such as a system persuading users in order to test the effects of alternative treatments or drugs to be sold by the system's company, or a financial planning or insurance software that persuades users to invest in the stock market to know the effect when the market crashes and leaves those users in financial ruin.

We are cognizant that, inevitably, we need to demonstrate the actual user-system interactions in real-time that are characterized by dialogue utterances in natural language. While our retrieval-based model generates persuader utterances in real-time by retrieving relevant strategies from the causal graph and converting them into corresponding utterances using a TF-IDF weighted bag-of-words representation, implementing, validating, and demonstrating the actual user interface is beyond the scope of this work. What we have accomplished is the creation of a pipeline that generates persuasive dialogue policies, which can be integrated into an intelligent user interface to effectively persuade users in real-world settings. We definitely expect that these policies, along with the causality-based counterfactual data generated, will require further refinement as the system interacts with actual users and acquires new information that validates the augmented data and possibly expand it further.

\section{Conclusion and Future Work}\label{conclusion}
We addressed the limitations of current persuasive dialogue systems, particularly their difficulty in adapting to user behavioral states and understanding the causal principles that dictate the response policies of both the persuadee and the persuader, as well as the resulting persuasion outcomes. Our novel approach integrates causal discovery (via GRaSP) to infer the causal structure of persuadee and persuader behaviors, which serves as inputs to counterfactual reasoning (via BiCoGAN). This structure, manifested through their utterances, optimizes persuasive policies and, consequently, maximizes persuasion outcomes. Combined with optimal policy learning (via
D3QN), our method outperforms the baselines. Thus, we have evidence to suggest that our approach can enhance a dialogue system’s persuasive ability through its generated counterfactual actions and user states. All of this confirms the effectiveness of leveraging causal discovery and counterfactual reasoning in optimizing persuasive interactions.

In the future, we plan to delve deeper into the various types of noise, both latent and observable, that could be estimated or modeled to better understand the aspects significantly influencing user-system interactions, such as cognitive biases, preferences, social norms, and environmental context, among others. Furthermore, several open issues need to be addressed. While our method captures causal dependencies in dialogue strategies, it does not yet account for user individuality. Personality traits, beliefs, and socioeconomic factors influence persuasion receptiveness, which necessitate user profiling for better personalization. Moreover, persuasion dynamics evolve as users adjust their responses based on experience, which requires adaptive learning mechanisms to track shifting causal relationships. Our approach also relies on annotated datasets, which may introduce biases. Future work should explore self-supervised learning and weakly supervised causal discovery to reduce annotation dependence. Incorporating non-verbal cues and multi-turn interactions could further enhance the model’s ability to optimize engagement and persuasion outcomes in real-world settings. Finally, this work does not include a real-world user study that would have provided insights into how actual users perceive the proposed dialogue mechanism. While our objective is to equip a persuasive agent with the intelligence derived from our personalized causality-based generative approach, it is essential for future work to develop an actual interface that enables persuasive system-user interactions.

\bibliographystyle{plain}
\bibliography{samplebib}

\begin{thebibliography}{10}

\bibitem{braca2023developing}
Annye Braca and Pierpaolo Dondio.
\newblock Developing persuasive systems for marketing: The interplay of persuasion techniques, customer traits and persuasive message design.
\newblock {\em Italian Journal of Marketing}, 2023(3):369--412, 2023.

\bibitem{chakrabarty2019}
Tuhin Chakrabarty, Christopher Hidey, Smaranda Muresan, Kathy McKeown, and Alyssa Hwang.
\newblock {AMPERSAND}: Argument mining for {PERS}u{A}sive o{N}line discussions.
\newblock In {\em EMNLP-IJCNLP}, pages 2933--2943, Hong Kong, China, November 2019. ACL.

\bibitem{chickering2002optimal}
David~Maxwell Chickering.
\newblock Optimal structure identification with greedy search.
\newblock {\em Journal of machine learning research}, 3(Nov):507--554, 2002.

\bibitem{devlin2018bert}
Jacob Devlin, Ming-Wei Chang, Kenton Lee, and Kristina Toutanova.
\newblock Bert: Pre-training of deep bidirectional transformers for language understanding.
\newblock {\em arXiv preprint arXiv:1810.04805}, 2018.

\bibitem{dillard2005nature}
James~Price Dillard and Lijiang Shen.
\newblock On the nature of reactance and its role in persuasive health communication.
\newblock {\em Communication monographs}, 72(2):144--168, 2005.

\bibitem{Furumai2024PersuaBot}
Kazuaki Furumai, Roberto Legaspi, Julio Vizcarra, Yudai Yamazaki, Yasutaka Nishimura, Sina~J. Semnani, Kazushi Ikeda, Weiyan Shi, and Monica~S. Lam.
\newblock Zero-shot persuasive chatbots with llm-generated strategies and information retrieval, 2024.

\bibitem{Hiraoka2016}
Takuya Hiraoka, Graham Neubig, Sakriani Sakti, Tomoki Toda, and Satoshi Nakamura.
\newblock Learning cooperative persuasive dialogue policies using framing.
\newblock {\em Speech Communication}, 84:83--96, 2016.

\bibitem{hirsh2012personalized}
Jacob~B Hirsh, Sonia~K Kang, and Galen~V Bodenhausen.
\newblock Personalized persuasion: Tailoring persuasive appeals to recipients’ personality traits.
\newblock {\em Psychological science}, 23(6):578--581, 2012.

\bibitem{hoch1985counterfactual}
Stephen~J Hoch.
\newblock Counterfactual reasoning and accuracy in predicting personal events.
\newblock {\em Journal of Experimental Psychology: Learning, Memory, and Cognition}, 11(4):719, 1985.

\bibitem{jaiswal2019bidirectional}
Ayush Jaiswal, Wael AbdAlmageed, Yue Wu, and Premkumar Natarajan.
\newblock Bidirectional conditional generative adversarial networks.
\newblock In {\em ACCV 2018: 14th Asian Conference on Computer Vision, Perth, Australia, December 2--6, 2018, Revised Selected Papers, Part III 14}, pages 216--232. Springer, 2019.

\bibitem{kaelbling1996reinforcement}
Leslie~Pack Kaelbling, Michael~L Littman, and Andrew~W Moore.
\newblock Reinforcement learning: A survey.
\newblock {\em Journal of artificial intelligence research}, 4:237--285, 1996.

\bibitem{kelders2012persuasive}
Saskia~M Kelders, Robin~N Kok, Hans~C Ossebaard, and Julia~EWC Van Gemert-Pijnen.
\newblock Persuasive system design does matter: a systematic review of adherence to web-based interventions.
\newblock {\em Journal of medical Internet research}, 14(6):e152, 2012.

\bibitem{lam2022greedy}
Wai-Yin Lam, Bryan Andrews, and Joseph Ramsey.
\newblock Greedy relaxations of the sparsest permutation algorithm.
\newblock In {\em Uncertainty in Artificial Intelligence}, pages 1052--1062. PMLR, 2022.

\bibitem{lee2006situation}
Cheongjae Lee, Sangkeun Jung, Jihyun Eun, Minwoo Jeong, and Gary~Geunbae Lee.
\newblock A situation-based dialogue management using dialogue examples.
\newblock In {\em 2006 IEEE International Conference on Acoustics Speech and Signal Processing Proceedings}, volume~1, pages I--I. IEEE, 2006.

\bibitem{lu2020sample}
Chaochao Lu, Biwei Huang, Ke~Wang, Jos{\'e}~Miguel Hern{\'a}ndez-Lobato, Kun Zhang, and Bernhard Sch{\"o}lkopf.
\newblock Sample-efficient reinforcement learning via counterfactual-based data augmentation.
\newblock {\em arXiv preprint arXiv:2012.09092}, 2020.

\bibitem{mnih2015human}
Volodymyr Mnih, Koray Kavukcuoglu, David Silver, Andrei~A Rusu, Joel Veness, Marc~G Bellemare, Alex Graves, Martin Riedmiller, Andreas~K Fidjeland, Georg Ostrovski, et~al.
\newblock Human-level control through deep reinforcement learning.
\newblock {\em nature}, 518(7540):529--533, 2015.

\bibitem{papangelis2022understanding}
Alexandros Papangelis, Nicole Chartier, Pankaj Rajan, Julia Hirschberg, and Dilek Hakkani-Tur.
\newblock Understanding how people rate their conversations.
\newblock In {\em Conversational AI for Natural Human-Centric Interaction: 12th International Workshop on Spoken Dialogue System Technology, IWSDS 2021, Singapore}, pages 179--189. Springer, 2022.

\bibitem{Pearl2009}
Judea Pearl.
\newblock {\em Causality: Models, Reasoning and Inference}.
\newblock Cambridge University Press, USA, 2nd edition, 2009.

\bibitem{pearl2000models}
Judea Pearl et~al.
\newblock Models, reasoning and inference.
\newblock {\em Cambridge, UK: CambridgeUniversityPress}, 19(2):3, 2000.

\bibitem{prakken2006formal}
Henry Prakken.
\newblock Formal systems for persuasion dialogue.
\newblock {\em The knowledge engineering review}, 21(2):163--188, 2006.

\bibitem{radford2019language}
Alec Radford, Jeffrey Wu, Rewon Child, David Luan, Dario Amodei, Ilya Sutskever, et~al.
\newblock Language models are unsupervised multitask learners.
\newblock {\em OpenAI blog}, 1(8):9, 2019.

\bibitem{raghu2017deep}
Aniruddh Raghu, Matthieu Komorowski, Imran Ahmed, Leo Celi, Peter Szolovits, and Marzyeh Ghassemi.
\newblock Deep reinforcement learning for sepsis treatment.
\newblock {\em arXiv preprint arXiv:1711.09602}, 2017.

\bibitem{ramsey2017million}
Joseph Ramsey, Madelyn Glymour, Ruben Sanchez-Romero, and Clark Glymour.
\newblock A million variables and more: the fast greedy equivalence search algorithm for learning high-dimensional graphical causal models, with an application to functional magnetic resonance images.
\newblock {\em International journal of data science and analytics}, 3:121--129, 2017.

\bibitem{reynolds2019psychological}
Tobias Reynolds-Tylus.
\newblock Psychological reactance and persuasive health communication: A review of the literature.
\newblock {\em Frontiers in Communication}, 4:56, 2019.

\bibitem{ribeiro2015influence}
Eug{\'e}nio Ribeiro, Ricardo Ribeiro, and David~Martins de~Matos.
\newblock The influence of context on dialogue act recognition.
\newblock {\em arXiv preprint arXiv:1506.00839}, 2015.

\bibitem{shi2020effects}
Weiyan Shi, Xuewei Wang, Yoo~Jung Oh, Jingwen Zhang, Saurav Sahay, and Zhou Yu.
\newblock Effects of persuasive dialogues: testing bot identities and inquiry strategies.
\newblock In {\em Proceedings of the 2020 CHI Conference on Human Factors in Computing Systems}, pages 1--13, 2020.

\bibitem{solus2021consistency}
Liam Solus, Yuhao Wang, and Caroline Uhler.
\newblock Consistency guarantees for greedy permutation-based causal inference algorithms.
\newblock {\em Biometrika}, 108(4):795--814, 2021.

\bibitem{spirtes2001causation}
Peter Spirtes, Clark Glymour, and Richard Scheines.
\newblock {\em Causation, prediction, and search}.
\newblock MIT press, 2001.

\bibitem{spirtes2013causal}
Peter~L Spirtes, Christopher Meek, and Thomas~S Richardson.
\newblock Causal inference in the presence of latent variables and selection bias.
\newblock {\em arXiv preprint arXiv:1302.4983}, 2013.

\bibitem{teyssier2012ordering}
Marc Teyssier and Daphne Koller.
\newblock Ordering-based search: A simple and effective algorithm for learning bayesian networks.
\newblock {\em arXiv preprint arXiv:1207.1429}, 2012.

\bibitem{Tiwari2022}
Abhisek Tiwari, Tulika Saha, Sriparna Saha, Shubhashis Sengupta, Anutosh Maitra, Roshni Ramnani, and Pushpak Bhattacharyya.
\newblock A persona aware persuasive dialogue policy for dynamic and co-operative goal setting.
\newblock {\em Expert Systems with Applications}, 195:116303.

\bibitem{torning2009persuasive}
Kristian Torning and Harri Oinas-Kukkonen.
\newblock Persuasive system design: state of the art and future directions.
\newblock In {\em Proceedings of the 4th international conference on persuasive technology}, pages 1--8, 2009.

\bibitem{tran2022ask}
Nhat Tran, Malihe Alikhani, and Diane Litman.
\newblock How to ask for donations? learning user-specific persuasive dialogue policies through online interactions.
\newblock In {\em Proceedings of the 30th ACM Conference on User Modeling, Adaptation and Personalization}, pages 12--22, 2022.

\bibitem{wachsmuth2017computational}
Henning Wachsmuth, Nona Naderi, Yufang Hou, Yonatan Bilu, Vinodkumar Prabhakaran, Tim~Alberdingk Thijm, Graeme Hirst, and Benno Stein.
\newblock Computational argumentation quality assessment in natural language.
\newblock In {\em Proceedings of the 15th Conference of the European Chapter of the Association for Computational Linguistics: Volume 1, Long Papers}, pages 176--187, 2017.

\bibitem{WangSKOYZY19}
Xuewei Wang, Weiyan Shi, Richard Kim, Yoojung Oh, Sijia Yang, Jingwen Zhang, and Zhou Yu.
\newblock Persuasion for good: Towards a personalized persuasive dialogue system for social good.
\newblock In Anna Korhonen, David~R. Traum, and Llu{\'{\i}}s M{\`{a}}rquez, editors, {\em ACL 2019}, pages 5635--5649. ACL, 2019.

\bibitem{weietal2016post}
Zhongyu Wei, Yang Liu, and Yi~Li.
\newblock Is this post persuasive? ranking argumentative comments in online forum.
\newblock In Katrin Erk and Noah~A. Smith, editors, {\em Proceedings of the 54th Annual Meeting of the Association for Computational Linguistics (Volume 2: Short Papers)}, pages 195--200, Berlin, Germany, August 2016. ACL.

\bibitem{wei2016post}
Zhongyu Wei, Yang Liu, and Yi~Li.
\newblock Is this post persuasive? ranking argumentative comments in online forum.
\newblock In {\em Proceedings of the 54th Annual Meeting of the Association for Computational Linguistics (Volume 2: Short Papers)}, pages 195--200, 2016.

\bibitem{yang2019let}
Diyi Yang, Jiaao Chen, Zichao Yang, Dan Jurafsky, and Eduard Hovy.
\newblock Let’s make your request more persuasive: Modeling persuasive strategies via semi-supervised neural nets on crowdfunding platforms.
\newblock In {\em Proceedings of the 2019 Conference of the North American Chapter of the Association for Computational Linguistics: Human Language Technologies, Volume 1 (Long and Short Papers)}, pages 3620--3630, 2019.

\bibitem{yoshino2018dialogue}
Koichiro Yoshino, Yoko Ishikawa, Masahiro Mizukami, Yu~Suzuki, Sakriani Sakti, and Satoshi Nakamura.
\newblock Dialogue scenario collection of persuasive dialogue with emotional expressions via crowdsourcing.
\newblock In {\em LREC 2018}, 2018.

\bibitem{zeng2024counterfactual}
Donghuo Zeng, Roberto~S Legaspi, Yuewen Sun, Xinshuai Dong, Kazushi Ikeda, Peter Spirtes, and Kun Zhang.
\newblock Counterfactual reasoning using predicted latent personality dimensions for optimizing persuasion outcome.
\newblock In {\em International Conference on Persuasive Technology}, pages 287--300. Springer, 2024.

\bibitem{zheng2024causal}
Yujia Zheng, Biwei Huang, Wei Chen, Joseph Ramsey, Mingming Gong, Ruichu Cai, Shohei Shimizu, Peter Spirtes, and Kun Zhang.
\newblock Causal-learn: Causal discovery in python.
\newblock {\em Journal of Machine Learning Research}, 25(60):1--8, 2024.

\end{thebibliography}

\end{document}